\documentclass{article}

\usepackage[final,nonatbib]{neurips_2021}

\usepackage[utf8]{inputenc} 
\usepackage[T1]{fontenc}    
\usepackage{hyperref}       
\usepackage{url}            
\usepackage{booktabs}       
\usepackage{amsfonts}       
\usepackage{nicefrac}       
\usepackage{microtype}      
\usepackage{xcolor}         

\usepackage{enumitem}
\usepackage[numbers]{natbib}
\usepackage{wrapfig}
\usepackage{caption}
\usepackage{subcaption}
\usepackage{graphicx}
\usepackage{amsmath}
\usepackage{dsfont}
\usepackage{verbatim}
\usepackage{mathabx}

\makeatletter
\let\@fnsymbol\@arabic
\makeatother

\newcommand*\diff{\mathop{}\!\mathrm{d}}

\usepackage{amsmath,amsfonts,bm}

\def\eqref#1{equation~\ref{#1}}

\def\1{\bm{1}}

\DeclareMathAlphabet{\mathsfit}{\encodingdefault}{\sfdefault}{m}{sl}
\SetMathAlphabet{\mathsfit}{bold}{\encodingdefault}{\sfdefault}{bx}{n}

\title{Score-Based Generative Classifiers}

\author{
  Roland S. Zimmermann\thanks{University of Tübingen, $^2$ Stanford University, $^3$ Norwegian University of Science and Technology.} \\
  \texttt{roland.zimmermann@uni-tuebingen.de} \\
  
  \And
  Lukas Schott$^1$ \\
  \texttt{lukas.schott@gmail.com} \\
  
  \And
  Yang Song$^2$ \\
  \texttt{yangsong@cs.stanford.edu} \\
  
  \And
  Benjamin A. Dunn$^3$ \\
  \texttt{benjamin.dunn@ntnu.no} \\
    
  \And
  David A. Klindt$^3$ \\
  \texttt{klindt.david@gmail.com} \\
}

\begin{document}

\maketitle

\begin{abstract}
    The tremendous success of generative models in recent years raises the question whether they can also be used to perform classification.
    Generative models have been used as adversarially robust classifiers on simple datasets such as MNIST, but this robustness has not been observed on more complex datasets like CIFAR-10.
    Additionally, on natural image datasets, previous results have suggested a trade-off between the likelihood of the data and classification accuracy.
    In this work, we investigate score-based generative models as classifiers for natural images.
    We show that these models not only obtain competitive likelihood values but simultaneously achieve state-of-the-art classification accuracy for generative classifiers on CIFAR-10.
    Nevertheless, we find that these models are only slightly, if at all, more robust than discriminative baseline models on out-of-distribution tasks based on common image corruptions.
    Similarly and contrary to prior results, we find that score-based are prone to worst-case distribution shifts in the form of adversarial perturbations.
    Our work highlights that score-based generative models are closing the gap in classification accuracy compared to standard discriminative models.
    While they do not yet deliver on the promise of adversarial and out-of-domain robustness, they provide a different approach to classification that warrants further research.
\end{abstract}

\section{Introduction}

\begin{wrapfigure}{R}{0.55\textwidth}
  \vspace{-25pt}
  \centering
  \includegraphics[width=0.47\textwidth]{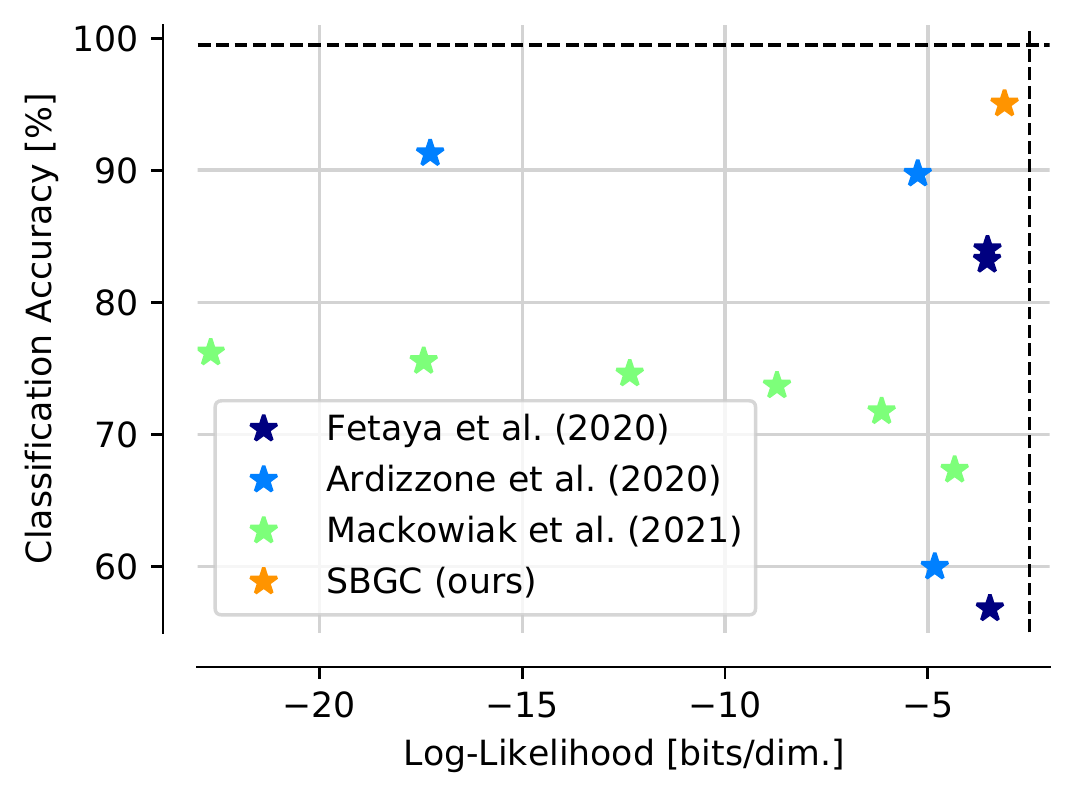} 
  \vspace{-5pt}
  \caption{\textbf{Model Comparison}. 
  Previous approaches \citep{fetaya2019understanding,ardizzone2020training,mackowiak2021generative} have demonstrated a trade-off between accuracy and likelihoods of generative classifiers on CIFAR-10. The black lines show current SOTA discriminative \citep[horizontal,][]{dosovitskiy2020image} and generative \citep[vertical,][]{kingma2021variational} models.
  }
   \label{fig:tradeoff}
  \vspace{-10pt}
\end{wrapfigure}

There exist two fundamentally distinct ways of performing classification.
The standard way is to train a classifier which discriminates between classes by modeling the conditional probability $p(y|\mathbf{x})$ of labels $y$ given some input $\mathbf{x}$.
These models are called \textit{discriminative} classifiers and have been extremely successful in supervised learning \citep{lecun1998gradient,krizhevsky2009learning}.
The alternative is to model the conditional likelihood $p(\mathbf{x}|y)$ of an image for each label and predict the label that maximizes this likelihood \citep{revow1996using}.
This approach is much less common because it requires solving the more complicated task of learning conditional generative models of the data; these models are referred to as \textit{generative} classifiers.

A line of research has been discussing the advantages and disadvantages between using discriminative or generative models for classification \citep{revow1996using,ng2002discriminative,raina2003classification,bouchard2004tradeoff,liang2008asymptotic}.
They found generative models to be more data efficient, but discriminative models to be asymptotically better.
However, these analyses were mostly limited to shallow models \citep[but see][]{goodfellow2013multi} and performance evaluation within the same data domain.
On MNIST \citep{lecun1998gradient} and SVHN \citep{netzer2011reading}, previous results have demonstrated that generative classifiers are robust even to worst-case distribution shifts in the form of adversarial perturbations \citep{schott2018towards,ju2020abs,paiton2020selectivity}.
Furthermore, it was found that generative classifiers have the highest alignment with humans on \emph{controversial} stimuli \citep{golan2020controversial}.

However, so far, these results have not transferred to any natural datasets of higher visual complexity \citep{li2019generative,fetaya2019understanding,ardizzone2020training,mackowiak2021generative}.
\citet{fetaya2019understanding} argue that `obtaining strong classification accuracy without harming likelihood estimation is still a challenging problem'.
This is empirically supported in their paper as well as a number of related works \citep{grathwohl2019your,ardizzone2020training,mackowiak2021generative} who all demonstrate a trade-off between good likelihoods and high classification accuracy, suggesting that to better capture what is unique for a label, some performance of the generative model needs to be sacrificed \citep{fetaya2019understanding}.
We challenge these ideas and add a new datapoint to the ongoing debate around generative classifiers demonstrating that near state-of-the-art likelihoods can be combined with classification accuracy at the level of discriminative models \citep{targ2016resnet}.

More specifically, score-based generative models have been proposed in recent years as a promising approach towards modeling complex high-dimensional data distributions \citep{sohl2015deep,ho2020denoising,song2020score}.
They are built on the idea of gradually diffusing a complex data distribution with a parameterized stochastic differential equation into a tractable noise distribution.
Previous work has suggested the effectiveness of these diffusion-based models for out-of-domain (OOD) tasks such as adversarially robust classification \citep{yoon2021adversarial} and as a general prior for natural images \citep{venkatakrishnan2013plug,kadkhodaie2020solving}.

In this work, we propose score-based generative classifiers (SBGC) and show that while these models overcome previous limitations --- such as trading off likelihood for accuracy --- they still do not solve the problem of worst-case distribution shifts on natural images.
More precisely, using our SBGC model, we report state-of-the-art classification accuracy for generative models on CIFAR-10 \citep{krizhevsky2009learning}, along with highly competitive likelihood values.
Nonetheless, we find that score-based generative classifiers show only slight improvements, if at all, in OOD benchmarks and are vulnerable to adversarial examples.

\section{Methods}

\subsection{Score-Based Generative Models}
This work builds on recent advances in score-based generative models of natural images \citep{sohl2015deep,ho2020denoising,song2020score,durkan2021maximum,kingma2021variational}.
The general idea behind this approach is to define a stochastic differential equation (SDE) describing the temporal transformation of the complex data distribution ($p$) into a tractable noise distribution ($p_T$, isotropic Gaussian).
This dynamical process can then be reversed by learning a time-dependent score function $\mathbf{s}_{\boldsymbol{\theta}}(\mathbf{x}(t), t)$ modeled by a neural network with parameters $\boldsymbol{\theta}$. 
Intuitively, this score function acts as an infinitesimal denoiser \citep{kadkhodaie2020solving}.
To generate new samples from the data distribution, one starts with random noise $\mathbf{x}(T)$ and follows the dynamics of the reverse SDE to produce a sample $\mathbf{x}(0)$ from the data distribution \citep[see][for more details]{song2020score}.

Moreover, by removing the stochasticity, the SDE becomes a (neural) ordinary differential equation \citep{song2020score,chen2018neural}.
Using a continuous time analog of the change of variables formula, one can compute the likelihood ($p_0$) of an input image $\mathbf{x}(0)$ under the model \citep{song2020score} as
\begin{align}\label{eq:uncon_llh}
    \log p_0(\mathbf{x}(0)) = \log p_T(\mathbf{x}(T)) + \int_0^T \nabla \cdot \widetilde{\mathbf{f}}_{\boldsymbol{\theta}}(\mathbf{x}(t), t) \,\diff t \\
    \text{with} \quad \widetilde{\mathbf{f}}_{\boldsymbol{\theta}}(\mathbf{x}(t), t) = \mathbf{f}(\mathbf{x}, t) - \frac{1}{2}g(t)^2\mathbf{s}_{\boldsymbol{\theta}}(\mathbf{x},t).
\end{align}
For the drift and diffusion coefficients $\mathbf{f}(\mathbf{x}, t)$ and $g(t)$, we used the same functions as the sub-VP model of \citet{song2020score}.
We use the training objective suggested in \citet{durkan2021maximum} which maximizes the likelihood of the training data under the model.
This is equivalent to minimizing the Kullback-Leibler divergence (KL) between the model and the data distribution $\min_\theta \operatorname{KL}(p_0|p)$ \citep{murphy2013machine}.

\subsection{Score-Based Generative Classifiers}
The most straightforward application of score-based generative models as classifiers would be to train one model per class as in \citet{schott2018towards}.
However, this produces very unstable training and considerable overfitting because the dataset size is, for each model, effectively reduced by one order of magnitude.
Instead, we add the image label $y \in \{1, ..., 10\}$ as a conditioning variable to the score function $f_\theta(\mathbf{x}(t), t, y)$; note that this also reduces the required amount of memory for multiple models.
In terms of the network architecture, we integrate this additional input by first converting it into a one-hot vector and then adding it to the model's representation in the same way as the time variable \citep[see][]{song2020score}. This is in line with concurrent work of \citet{nichol2021improved} and loosely inspired by \citet{jun2020distribution}. 
For classification of a given input $\mathbf{x}(0)$, we can then condition the score function on each label $y$
\begin{equation}\label{eq:llh}
    \log p_{0}(\mathbf{x}(0)|y) = \log p_{T}\left(
    \mathbf{x}(0) + \int_0^T \widetilde{\mathbf{f}}_{\boldsymbol{\theta}}(\mathbf{x}(t), t, y) \diff t
    \right) + \int_0^T \nabla \cdot \widetilde{\mathbf{f}}_{\boldsymbol{\theta}}(\mathbf{x}(t), t, y) \diff t,
\end{equation}
where $\widetilde{\mathbf{f}}_{\boldsymbol{\theta}}$ is defined as above in Eq.~(\ref{eq:uncon_llh}). For any input image $\mathbf{x}$, the model then predicts the label that yields the highest conditional likelihood $\max_{y} \log p_{0}(\mathbf{x}(0)|y)$. To get the unconditional likelihood of samples we marginalize over the classes, i.e. $p(\mathbf{x}) = \sum_{y_i} p_0(\mathbf{x}|y=y_i)$.

\section{Results}

\subsection{Classification Accuracy and Data Likelihood}

\begin{table}[tb]
     \centering
     \caption{\textbf{Model Comparison.} Accuracy (in \%) and negative log-likelihoods (NLL) in bits per dimension \citep{theis2015note} on the CIFAR-10 test set.
     The lower half of the table shows a baseline discriminative model \cite{targ2016resnet} and the current state-of-the-art discriminative \cite{dosovitskiy2020image} and generative \cite{kingma2021variational} models.
     }
     \vspace{0.2cm}
     {\begin{tabular}{lcc}
     Model approach & Accuracy [\%] $\uparrow$ & NLL [bits/dim.] $\downarrow$ \\
     \midrule
     Invertible Network (\citet{mackowiak2021generative}) & 67.30 & 4.34 \\
     GLOW (\citet{fetaya2019understanding}) & 84.00 & 3.53 \\
     Normalizing flow (\citet{ardizzone2020training}) & 89.73 & 5.25 \\
     Energy model (\citet{grathwohl2019your}) & 92.90 & N/A \\
     SBGC (ours) & \textbf{95.04} & \textbf{3.11} \\
     \midrule
     WideResNet-28-12 (\citet{targ2016resnet}) & 95.42 & N/A \\
     ViT-H/14 (\citet{dosovitskiy2020image}) & \textbf{99.50} & N/A \\
     VDM (\citet{kingma2021variational}) & N/A & \textbf{2.49}
     \end{tabular}}
     \label{table:model_comparison}
\end{table}

First of all, we note that previous methods \citep{fetaya2019understanding,ardizzone2020training,mackowiak2021generative} have discussed a trade-off between classification accuracy and likelihoods (see Figure~\ref{fig:tradeoff}).
This is further supported by observations on MNIST, where increasing the latent dimensionality of a variational autoencoder increases the likelihood at the cost of decreasing classification accuracy and robustness \citep{chen2020breaking}.
Older work has focused on hybrid models that combine the data efficiency of (shallow) generative models with the asymptotic performance of discriminative models.
However, this comes with a cost in model likelihood \citep{fujino2005hybrid,raina2003classification}.

In contrast to these previous methods, we find that our model achieves both state-of-the-art accuracy and likelihoods for generative classifiers on CIFAR-10 (Table~\ref{table:model_comparison}).
We also test the deeper architecture proposed in \citet{durkan2021maximum}: While this does not change the classification accuracy, it slightly increases the NLL to $3.08$\, bits/dim. 
We leave further exploration of this direction to future research.

\subsection{Out-of-Distribution Robustness: Common Image Corruptions}

\begin{table}[htb]
     \centering
     \caption{\textbf{Performance on Common Image Corruptions.} Accuracy (in \%) on clean CIFAR-10 test set and the mean accuracy on CIFAR-10-C (considering a random subset with 10\% of the original size for SBGC because of computational limitations). We refer to random flips, random crops and uniform noise as simple augmentations.
     }
     \vspace{0.2cm}
     {\begin{tabular}{lccc}
     & & \multicolumn{2}{c}{CIFAR-10-C} \\
     Models w/ data augm. & CIFAR-10 & all corruptions & w/o noises \\
     \midrule
     ResNeXt29 + AugMix (\citet{hendrycks2019augmix}) & \textbf{95.83} & 89.09 & 90.51 \\ 	
     ResNet-50 + adv. augm. (\citet{calian2021defending}) & 94.93 & \textbf{92.17} & \textbf{92.53}\\
     \\
     Models w/ simple data augm.\\
     \midrule
     %
     WideResNet-28-12 (\citet{targ2016resnet}) & \textbf{95.42} & 74.42 & \textbf{80.21} \\
     SBGC (ours) & 95.04 & \textbf{76.24} & 75.71 \\
     \end{tabular}}
     \label{table:model_comparison_C}
\end{table}

Most previous approaches towards improving out-of-distribution performance on common image corruptions \citep{hendrycks2019benchmarking} have relied on some form of data augmentation in the form of carefully hand-crafted transformations \citep{hendrycks2019augmix} or adversarial training \citep{calian2021defending}.
Here, we propose an orthogonal approach that is based on different modeling assumptions.
Specifically, we are interested in seeing whether the inductive bias implicit in score-based generative classifiers improves their classification accuracy when generalizing to the image domains of CIFAR-10-C \citep{hendrycks2019benchmarking}.
We find that our model performs better than some previous models that are solely trained on the original data (Table~\ref{table:model_comparison_C}). 

Since score-based models are trained like denoisers, they might work specifically well on noisy data. 
Thus, just comparing the mean accuracy over all corruption types, could give them an unfair advantage over the baselines.
Hence, we also compare the mean accuracy for all corruptions excluding noises \footnote{Gaussian noise, shot nose, impulse noise} (Table~\ref{table:model_comparison_C}). Here, we see that our SBGC model does not perform better than the baselines.
We conclude that the improvements of the SBGC model in terms of out-of-distribution robustness are limited to noise corruptions.

When integrating data augmentation methods, discriminative models achieve higher performance (Table~\ref{table:model_comparison_C}, top).
However, we note that the approach of augmenting datasets is independent of our suggested changes, with the caveat that changing the dataset needs to be done carefully for generative models \citep{jun2020distribution}.
Preliminary experiments indicate that we can increase our performance on CIFAR-10-C further by leveraging more data augmentations \citep{hendrycks2019augmix} (accuracy $\sim83\%$).
However, there is still a large gap towards discriminative models with data augmentation.

\subsection{Out-of-Distribution Robustness: Adversarial Perturbations}
Despite their small magnitude, adversarial perturbations can cause a drastic change in the behavior of a neural network \citep{szegedy2013intriguing} and can be seen as a worst-case distribution shift.
To assess the robustness of our SBGC model against such perturbations, we first generate them using a projected gradient descent (PGD) \citep{madry2018towards} on the negative cross-entropy of the model's prediction using Foolbox \citep{rauber2020foolbox}. 
For this, we estimate the gradients of the model likelihoods with respect to the inputs $\diff p_{0,y}(\mathbf{x}) / \diff \mathbf{x}$ using the adjoint sensitivity method \citep{chen2018neural}.
This requires second-order optimization because of the Jacobian term in equation~\ref{eq:llh}.
We implement this efficiently with Hessian-vector-products in JAX \citep{jax2018github} and verify the implementation by comparing the calculated gradient with numerical estimations based on finite differences.
Since computing gradients is still computationally expensive (a single forward and backward pass take approximately 90 minutes on an NVIDIA Tesla V100 GPU), we follow common practice in this domain and perform the adversarial evaluation on a reduced sample size of $N = 216$.

While our approach partially improves robustness against common image corruptions, it strikingly fails in the presence of adversarial perturbations (Tab.~\ref{tab:adversarial_robustness}).
Within both standard $\ell_\infty$ perturbation norms ($8/255$) as well as standard $\ell_2$ perturbation norms ($0.5$), our model miss-classifies every adversarially perturbed input image.

\begin{table}[htb]
    \centering
    \caption{\textbf{Performance on Adversarial Perturbations.} Accuracy (in \%) of different models against adversarial perturbations generated by a norm-bounded $\ell_\infty$ and $\ell_2$ PGD \citep{madry2018towards} attack (against each specific model) with a bound of $\epsilon=8/255$ and $\epsilon=0.5$, respectively.
    The bottom row shows the current state-of-the-art model for adversarially robust classification on CIFAR-10 as a reference value.
    }
    \vspace{0.2cm}
    \begin{tabular}{lccc}
        Model & Clean & $\ell_\infty$ \ & $\ell_2$ \\
        \toprule
        SBGC (ours) & \textbf{95.04} & $0.00$ & $0.00$ \\
        WideResNet-70-16 + augm. (\citet{rebuffi2021fixing}) & 92.23 & \textbf{82.32} & \textbf{66.56} \\
    \end{tabular}
    \label{tab:adversarial_robustness}
\end{table}

\subsection{Likelihood Estimation on Interpolated Samples}

\begin{figure}[htb]
  \centering
  \includegraphics[scale=0.7]{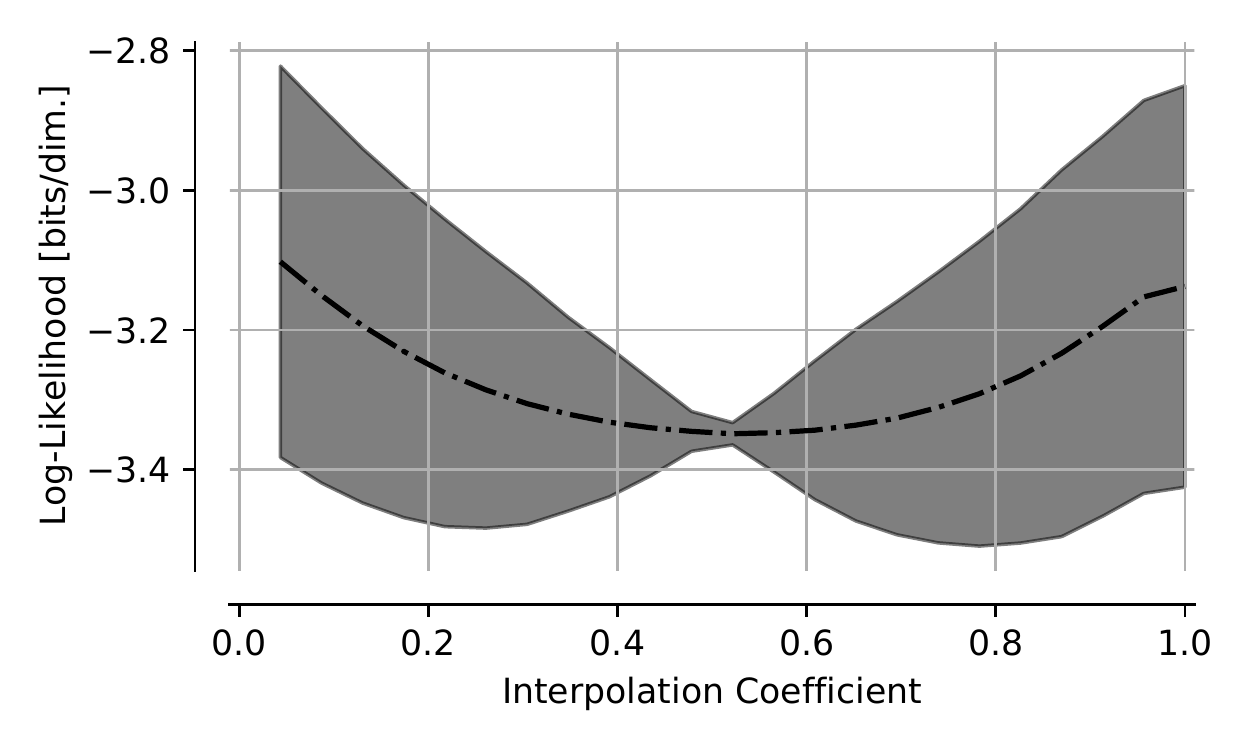} 
  \caption{\textbf{Interpolation Experiment}. We linearly interpolate between dataset samples and asses the shape of the log-likelihood interpolation function (equation \ref{eq:interpolation}). While \citet{fetaya2019understanding} found a concave shape, indicating high likelihoods outside the data distribution, we find a convex function suggesting that the score based generative model does assign lower likelihoods outside the training data distribution. The black line shows the mean and the shaded area indicates the standard deviation of the estimated log-likelihood over $500$ image pairs.}
   \label{fig:interpolation_experiment}
\end{figure}

In a previous study, \citet{fetaya2019understanding} highlighted a problem with generative classifiers.
They found that the likelihood of an interpolated image on CIFAR-10 is actually higher than the likelihoods of both the start $x_A$ and endpoint $x_B$ of the interpolation.
More precisely, they demonstrated that the function
\begin{equation}\label{eq:interpolation}
    f(t) = p_0\left(t \cdot x_A + (1-t) \cdot x_B\right), \quad t \in [0, 1]
\end{equation}
is concave, which is counter-intuitive because a generative model should assign low likelihood to inputs outside of its training distribution such as nonsensical interpolations in pixel space.

To test this for our SBGC model, we measure the function $f(t)$ for $24$ values for $t \in [0, 1]$ and average this over $500$ image pairs. As depicted in Figure~\ref{fig:interpolation_experiment}, we find that the interpolation of the likelihood is generally convex.
Thus, in contrast to the analysis of previous models by \citep{fetaya2019understanding}, our SBGC model assigns lower likelihood values to interpolated images that are outside the (training) data distribution.

\section{Conclusion}
We have shown that the latest advances in score-based generative modeling of natural images translate into generative classifiers which have highly competitive classification accuracies as well as likelihoods.
However, in the search for further benefits, we found that these models show only minor or no improvements on common image corruptions.
Similarly, but in contrast to previous results \citep{yoon2021adversarial}, they spectacularly fail on worst-case distribution shifts.
This highlights the necessity to perform extensive and, if at all possible, gradient-based adversarial attacks.
The lack of adversarial robustness aligns with previous results \citep{kos2018adversarial,li2019generative,ardizzone2020training,mackowiak2021generative}.
Interestingly, this vulnerability remains \emph{despite} resolving the problem of high likelihood on interpolated images \citep{fetaya2019understanding}, and so we can exclude this as a possible explanation for the lacking robustness. 

Thus, we are left wondering why generative classification of natural images does not show the same robustness to worst-case distribution shifts as observed on MNIST \citep{schott2018towards}.
A possible explanation for the positive results on MNIST might be that Euclidean distances (on which the likelihood of variational autoencoders with a Gaussian likelihood is based) are a useful metric on MNIST but not on real images.
This also led to the provable robustness on MNIST \citep{schott2018towards}.
However, on natural images Euclidean distances are badly aligned with human perception \citep{teo1994perceptual,wang2004image,laparra2016perceptual} and we have no guarantee about the model behavior even in small $\ell_p$ epsilon balls around the training data.
Further, the results of \citet{chen2020breaking} offer an additional explanation as they hypothesize that a low dimensional latent space is a necessary property of a robust generative classifier. Since the latent space of our SBGC models has the same dimensionality as the images - and hence cannot be described as low dimensional - this might explain the lack of robustness to worst-case distribution shifts.

In a nutshell, while we achieve almost the same accuracy on clean data as a baseline discriminative classifier and show high data likelihoods, we find no advantage over discriminative classifiers regarding worst-case distribution shifts and common corruptions.

\section*{Acknowledgments and Funding Sources}
We thank Matthias Bethge, Wieland Brendel, Will Grathwohl, Zahra Kadkhodaie, Dylan Paiton, Ben Poole, Yash Sharma and Eero Simoncelli for their feedback.
Further, we thank the International Max Planck Research School for Intelligent Systems (IMPRS-IS) for supporting RSZ. This work was partially supported by a Research Council of Norway FRIPRO grant (90532703) and the German Federal Ministry of Education and Research (BMBF) through the Competence Center for Machine Learning (TUE.AI, FKZ 01IS18039A).

\bibliography{references}
\bibliographystyle{plainnatnourl}

\newpage
\appendix
\section{Appendix}

\subsection{Training and Implementation Details}

We build on the architecture (baseline model with $4$ residual blocks) proposed by \citet{durkan2021maximum} and extend it to condition the score function also on the class label. All in all, our model has $67.19$\,M parameters. Furthermore, we use the same SDE (variance preserving) and maximum likelihood training procedure (trained for $950,000$ steps with a batch size of $128$ and importance sampling) as \citet{durkan2021maximum}.
The training takes approximately five days on two NVIDIA Tesla V100 GPUs.

For the likelihood inference, i.e., for solving the SDE, we use the Runge-Kutta 4(5) solver of SciPy (\emph{scipy.integrate.solve\_ivp}) \citep{2020SciPy-NMeth} with absolute and relative precision equal to $10^{-5}$.
We perform uniform dequantization (i.e., adding uniform noise with a magnitude of $1/256$) and integrate the SDE up to a minimal $\epsilon=10^{-5}$ for numerical stability \citep{song2020score}.
Furthermore, the final likelihood is computed by taking the average over $n=30$ random draws for the epsilon trace estimator \citep{song2020score}.
For the OOD tasks (CIFAR-10-C and adversarial perturbations), we lower this to $n=10$ to reduce the compute time, thus giving a more conservative lower estimate of the performance on these datasets (see also section~\ref{appendix:additional_analyses_convergence}).

For the WideResNet-28-12 baseline we use an existing implementation\footnote{\url{https://github.com/meliketoy/wide-resnet.pytorch}} in PyTorch. To make this baseline more comparable with our SBGC, we also use dequantization noise for augmenting the training data of the WideResNet.

\subsection{Additional Analyses}\label{appendix:additional_analyses}

\subsubsection{Convergence Analysis}\label{appendix:additional_analyses_convergence}

\begin{figure}[htb]
  \makebox[\textwidth][c]{\includegraphics[width=1.0\textwidth]{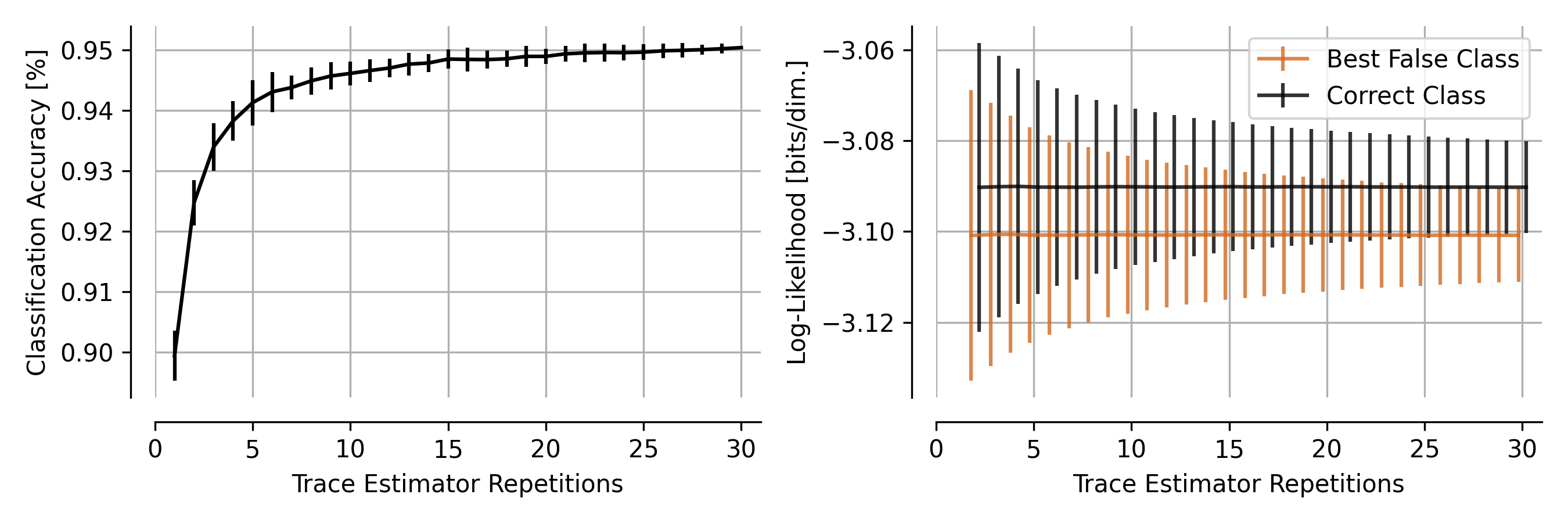}}
  \caption{\textbf{Convergence Analysis.} The left plot shows the classification accuracy as a function of the number of repetitions for the trace estimator (error bars indicate two standard deviations over random subsamples).
  The right plot shows the estimate of the log-likelihood as a function of the number of repetitions for the trace estimator (error bars indicate standard error of the mean) for both the correct class (black) as well the wrong class with the highest likelihood (orange).}
  \label{fig:convergence}
\end{figure}

Here, we study the effect of varying the number of repetitions of the trace estimator (see previous section).
In Figure.~\ref{fig:convergence} on the left we see that the classification accuracy improves monotonically with the number of repetitions.
A number of $n=10$ gives nearly asymptotic performance and is, thus, a good trade-off for our OOD tasks which are computationally intensive.
For asymptotic performance, we see that the accuracy rises above $95\%$.

On the right side of Figure.~\ref{fig:convergence}, we study the convergence of the trace estimator which is part of the likelihood (equation \ref{eq:llh}.
We can see that the standard error of the mean (errorbars) converges very slowly.
We therefore report the likelihood at $n=30$ repetitions.
Note also that the likelihood of the correct class converges (averaged across all test images) to a higher value as that of the wrong class with the highest likelihood.
This is expected for a model with good classification accuracy.
Thus, provided labels, our model attains slightly higher likelihoods than in the unconditioned case.

\begin{figure}[htb]
  \makebox[\textwidth][c]{\includegraphics[width=1.0\textwidth]{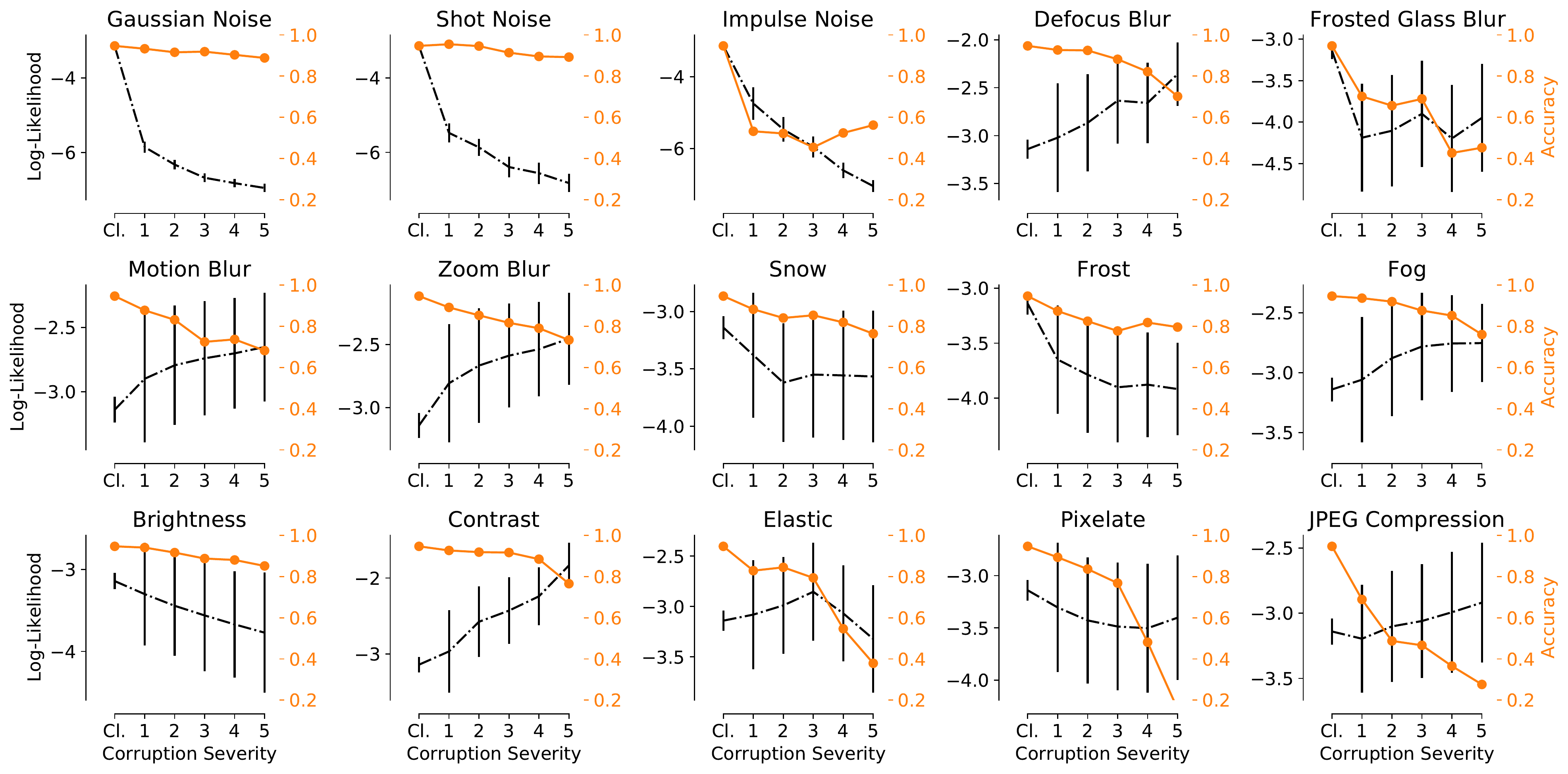}}%
  \caption{\textbf{Detailed Results on CIFAR-10-C.} We show the log-likelihood (orange, left axis) and the accuracy (black, right axis) across all image corruption types and severities (horizontal axis) for the CIFAR-10-C benchmark \citep{hendrycks2019benchmarking}. Note that the severity level \emph{Cl.} corresponds to clean samples.}
  \label{fig:cifar_c}
\end{figure}

\subsubsection{CIFAR-10-C Performance}
Here, we further resolve the model performance on CIFAR-10-C for all different corruption types and strengths in Figure~\ref{fig:cifar_c}.
Interestingly, the results highlight that likelihood and accuracy need not always be aligned.
For the different corruptions based on blurring or some sort of low-pass filtering (e.g., fog or JPEG compression), we observe that while the accuracy decreases (as signal is lost) the likelihood under the model actually increases.

This observation is in line with previous studies \citep{nalisnick2018deep}.
We hypothesize that this highlights the fact that natural images follow a $1/f$ power spectrum \citep{field1987relations} and, thus, low-pass filtering will increase the likelihood of any image.
This also explains previous observations of non-natural images obtaining high likelihoods under generative models:
Starting from a non-natural image we can always blur the image to obtain high likelihoods under our model (see, e.g., our Figure~\ref{fig:cifar_c} zoom blur and Table 3 in \citep{fetaya2019understanding}).
Similarly, we can start from a natural image (e.g., from the training set) and add white noise.
The high-frequency perturbations will quickly deteriorate the likelihoods (see Figure~\ref{fig:cifar_c} Gaussian noise).
Relatedly, \citet{nalisnick2019detecting} found that a measure of dataset complexity can be leveraged in conjunction with likelihood estimates to perform OOD detection.

\end{document}